\documentclass[runningheads]{llncs}
\usepackage{graphicx}
\usepackage{amsmath}
\usepackage{amsfonts}
\usepackage{tikz}
\usepackage{forest}
\usepackage{hf-tikz}
\usetikzlibrary{tikzmark}
\usepackage[ruled, linesnumbered]{algorithm2e}
\usepackage{algpseudocode}
\usepackage{subcaption}
\usepackage{booktabs}
\usepackage{pgfplots}
\usepgfplotslibrary{fillbetween}
\usepgfplotslibrary{groupplots}
\usepackage{filecontents}

\DeclareMathOperator*{\argmin}{arg\,min}
\DeclareMathOperator*{\mean}{mean}

\SetKwProg{Fn}{Function}{:}{end}
\SetFuncSty{textnormal}
\SetKwComment{Comment}{$\triangleright$\ }{}

\SetCommentSty{mycommfont}

\begin{document}
\title{Hierarchical Collaborative Hyper-parameter Tuning}
%
%
\author{Ahmad Esmaeili\inst{1}\orcidID{0000-0003-0612-2351} \and
Zahra Ghorrati\inst{1} \and
Eric T. Matson\inst{1}}
\authorrunning{A. Esmaeili et al.}
%
\institute{Department of Computer and Information Technology, Purdue University, \\West Lafayette IN 47907, USA\\ 
\email{\{aesmaei, zghorrat, ematson\}@purdue.edu}\\
}
\maketitle              
\begin{abstract}
Hyper-parameter Tuning is among the most critical stages in building machine learning solutions. This paper demonstrates how multi-agent systems can be utilized to develop a distributed technique for determining near-optimal values for any arbitrary set of hyper-parameters in a machine learning model. The proposed method employs a distributedly formed hierarchical agent-based architecture for the cooperative searching procedure of tuning hyper-parameter values. The presented generic model is used to develop a guided randomized agent-based tuning technique, and its behavior is investigated in both machine learning and global function optimization applications. According the empirical results, the proposed model outperformed both of its underlying randomized tuning strategies in terms of classification error and function evaluations, notably in higher number of dimensions. 

\end{abstract}
\section{Introduction}
Almost all Machine Learning (ML) algorithms comprise a set of hyper-parameters that control their learning experience and the quality of their resulting models. The number of hidden units, the learning rate, the mini-batch sizes, etc. of a neural network; the kernel parameters and regularization penalty amount of a support vector machine; and maximum depth, samples split criteria, and the number of used features of a decision tree are few examples of hyper-parameters that need to be adjusted for the aforementioned learning methods. Numerous research studies have been devoted to hyper-parameter tuning in the machine learning literature. In the most straight forward manual approach an expert knowledge is used to identify and evaluate a set of potential values in the hyper-parameter search space. Accessing the expert knowledge and generating reproducible results are among the primary difficulties in using the manual searching technique \cite{bergstra2012random}, especially knowing that different datasets will likely require different set of hyper-parameter values for a specific learning model~\cite{kohavi1995automatic}.

Let $\Lambda=\{\lambda\}$ and $\mathcal{X}=\{\mathcal{X}^{(train)}, \mathcal{X}^{(valid)}\}$ respectively denote the set of all possible hyper-parameter value vectors and the data set split into training and validation sets. The learning algorithm with hyper-parameter values vector $\lambda$ is a function that maps training datasets to model $\mathcal{M}$, i.e. $\mathcal{M}=\mathcal{A}_\lambda(\mathcal{X}^{(train)})$, and the hyper-parameter optimization problem can be formally written as \cite{bergstra2012random}:
\begin{equation}\label{eq:HOP}
	\lambda^{(*)}=\argmin_{\lambda\in\Lambda}\mathbb{E}_{x\sim\mathcal{G}_x}\left[\mathcal{L}\left(x; \mathcal{A}_\lambda(\mathcal{X}^{(train)})\right)  \right] 
\end{equation}
where $\mathcal{G}_x$ is the grand truth distribution, $\mathcal{L}(x;\mathcal{M})$ is the expected loss of applying learning model $\mathcal{M}$ over i.i.d. samples $x$, and $\mathbb{E}_{x\sim\mathcal{G}_x}\left[\mathcal{L}\left(x; \mathcal{A}_\lambda(\mathcal{X}^{(train)})\right)  \right]$ is the generalization error of algorithm $\mathcal{A}_\lambda$. Estimating the generalization error using the cross-validation technique \cite{10.1162/EVCO_a_00069}, the above-mentioned optimization problem can be rewritten as:
\begin{equation}\label{eq:HOP_CV}
	\lambda^{(*)}\approx\argmin_{\lambda\in\Lambda}\mean_{x\in\mathcal{X}^{(valid)}}\mathcal{L}\left(x; \mathcal{A}_\lambda(\mathcal{X}^{(train)})\right)\equiv\argmin_{\lambda\in\Lambda} \Psi(\lambda)
\end{equation}
where $\Psi(\lambda)$ is called hyper-parameter response function \cite{bergstra2012random}.

There are a broad range of methods that address this problem using global optimization techniques. Grid search \cite{montgomery2017design,John94cross-validatedc4.5:}, random search \cite{bergstra2012random}, Bayesian optimization \cite{movckus1975bayesian,mockus2012bayesian,feurer2019hyperparameter}, and evolutionary and population-based optimization \cite{simon2013evolutionary,esmaeili2009adjusting,9504761} are among the widely used class of approaches that are studied extensively in the literature. 
In grid search, for instance, every combination of predetermined set of values in each hyper-parameter is examined and the best point, i.e. the one that minimizes the loss, is selected. Assuming $\mathcal{V}_j$ and $k$ denote the set of candidate values for the hyper-parameter $\lambda_j^{(i)}\in \lambda^{(i)}$ and the number of hyper-parameter to configure, respectively, the number of trials that grid search evaluates is $T=\Pi_{j=1}^k|\mathcal{V}_j|$, which clearly indicates the curse of dimensionality because of the exponential growth of joint-values with the increase in the number of hyper-parameters \cite{bellman1961adaptive}. 

When applied to machine learning and data mining, Multi-Agent Systems (MAS) and agent-based technologies offer various benefits, such as scalability, facilitating the autonomy and distributedness of learning resources, and supporting strategic and collaborative learning models, to name a few \cite{ryzko2020modern,esmaeili2020hamlet}. Collaborative and agent-based methods to tune the hyper-parameters of a machine learning model has previously studied in the literature. Among the noteworthy contributions, the research reported in \cite{bardenet2013collaborative} proposes a surrogate-based collaborative tuning method that incorporates the gained experience from previous experiments. The collaboration in this model is basically through simultaneous tuning of same hyper-parameter sets over multiple datasets and using the obtained information about the optimized hyper-parameter values in all subsequent tuning problems. In \cite{swearingen2017atm}, a distributed and collaborative system called Auto-Tuned Models (ATM) is proposed to automate hyper-parameter tuning and classification model selection procedures. ATM uses Conditional Parameter Tree (CPT) to represent the hyper-parameter search space in which the root is learning method and the children of the root are hyper-parameters. During model selection, different tunable subsets of hyper-parameter nodes in CPT are selected and assigned to a cluster of workers to tune. Koch et al. \cite{koch2018autotune} has introduced a derivative-free hyper-parameter optimization framework called Autotune. This framework is composed of a hybrid and extendible set of solvers that concurrently run various searching methods, and a potentially distributed set of workers for evaluating objective functions and providing feedback to the solvers. Autotune uses an iterative process during which the solver manager exchanges all the points that have been evaluated with the solvers to generate new sets of points to evaluate.  
Iranfar et al. \cite{iranfar2021multi} have proposed a Mutli-Agent Reinforcement Learning (MARL)-based technique to optimize the hyper-parameters of deep convolutional neural networks. In their work, the design-space is split into sub-spaces by devoting each agent to tuning the hyper-parameters of a single network layer using Q-learning. In \cite{parker2020provably} Parker-Holder et al. introduce Population-Based Bandit (PB2) algorithm that utilizes a probabilistic model to efficiently guide the searching operation of hyper-parameters in Reinforcement Learning. In PB2, a population of agents are trained in parallel, and periodically observing their performance, the network weights of an under-performing agent is replaced with the ones of a better performing agent, and its hyper-parameters are tuned using Bayesian optimization.

Our paper presents a distributed agent-based collaborative optimization model that can be used to solve problems related to ML hyper-parameter tuning as well as generic function optimization. In the suggested method, in which each terminal agent focuses on sub-optimizing based on a single hyper-parameter, and the high-level agents are responsible for aggregating sub-optimal results and facilitating subordinates' collaborations. Thanks to the intrinsic autonomy of the agents in a MAS, each agent in the proposed model can flexibly utilize a different searching/optimization procedure to locate optima. Notwithstanding, for the sake of experiments, we have developed a guided random-based searching approach for all terminal agents, and have applied set theoretic operations as the aggregation techniques in hight-level agents. The suggested agent-based modeling of the problem enables the tuning process to be carried out based on multiple metrics and be easily implemented over not only parallel processing units in a single machine, but also multiple heterogeneous devices connected by a network.

The remainder of this paper is organized as follows: section~\ref{sec:proposed} presents the details of the proposed agent-based method, section~\ref{sec:results} discusses the performance of the proposed model and presents empirical results from applying it to ML hyper-parameter tuning and function optimization use cases, and at the end, section~\ref{sec:conclusion} concludes the paper and provides future work suggestions. 

\section{Methodology}\label{sec:proposed}
This section provides the details of the proposed agent-based hyper-parameter tuning approach. The first part presents the general algorithms for building the initial multi-agent structure and using it for tuning the hyper-parameters, and the second part uses the presented foundations to develop a guided randomized optimization and hyper-parameter tuning method.

\subsection{Agent-based Hyper-parameter Tuning}
The agents in our proposed model are basically the solvers that are specialized in tuning a specific subset of hyper-parameters. Because of the hierarchical representation of the model, we have two sets of agents: (1) \emph{internals}, which have subordinates and mainly focus on aggregating results and directing information; (2) \emph{terminals}, which do not have any child agents, and implement a single hyper-parameter tuning algorithm based on the provided information. Throughout this paper, we use notation $\mathcal{T}^{l}_{\lambda}$ to refer to the agent(tuner) that is at level $l$ of the hierarchy and specialized in tuning hyper-parameter set $\lambda$. 

\begin{figure}
	\centering
	\includegraphics[]{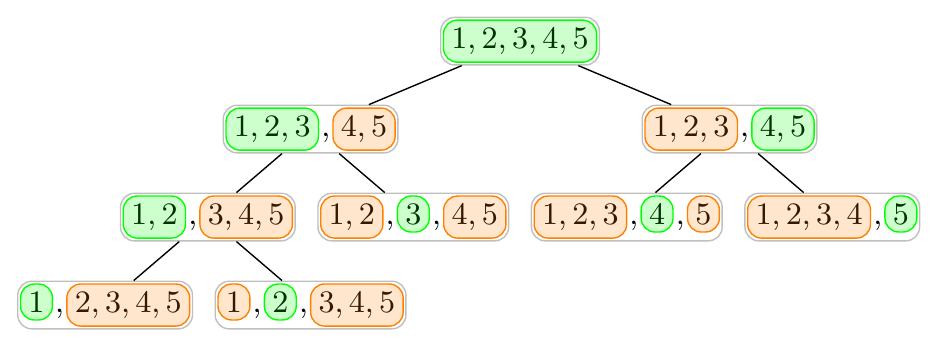}
	\caption{Hierarchical structure built for $\lambda_o=\{\lambda_1, \lambda_2, \lambda_3, \lambda_4, \lambda_5\}$, where the primary and complementary hyper-parameters of each node are respectively highlighted in green and orange, and the labels are the indexes of $\lambda_i$.}
	\label{fig:hyper-hier}
\end{figure}

The placement of agents in the hierarchy is determined by the hyper-parameter sets that they tune. Let $\lambda=\{\lambda_1, \lambda_2, \ldots, \lambda_n \}$ be the set of $n$ hyper-parameters used by ML algorithm $\mathcal{A}_\lambda$. We assume $\lambda=\lambda_o\cup\lambda_f$, where $\lambda_o$ is the sets of objective hyper-parameters we try to tune, and $\lambda_f$ denote the set of hyper-parameters we intend to keep their values fixed. We form the hierarchy of the agents by recursively dividing $\lambda_o$ into \emph{primary} hyper-parameter subsets, $\hat{\lambda}_o$, and assigning each, together with its \emph{complement}, $\hat{\lambda}'_o=\lambda_o-\hat{\lambda}_o$, to an agent. The recursive division is based on the maximum number of parallel connections, $c>1$, the agents can handle and continues until the hyper-parameter set is no longer divisible, i.e. $|\hat{\lambda}_o|=1$, where $|\ldots|$ denotes the set cardinality operator. Fig~\ref{fig:hyper-hier} illustrates an example structure built for $\lambda_o=\{\lambda_1, \lambda_2, \lambda_3, \lambda_4, \lambda_5\}$ and $c=2$. In this figure, green and orange highlighters are used to distinguish the primary and complementary hyper-parameters at each node respectively. Moreover, for the sake of brevity, we have used the indexes as the label of the nodes.

The proposed tuning technique is a two-phase process: (1) distributedly forming the hierarchical multi-agent system, and (2) iteratively tuning the hyper-parameters through the cooperation between the agents, both presented in algorithm~\ref{alg:form-hier}. Building the hierarchy is initiated when the root agent receives a tuning query made based on a user's request. Such query is characterized by tuple 
\begin{equation}
	\left<\mathcal{A}_\lambda,\{\hat{\lambda}_o,\hat{\lambda}'_o, \lambda_f\}, \mathcal{V}, \{\mathcal{X}^{(train)},\mathcal{X}^{(valid)}\}, \mathcal{L}\right>
\end{equation}
 where $\mathcal{V}=\{(\lambda_i, v_i)\}; |\mathcal{V}|\geq|\lambda_f|$ denotes the set containing  hyper-parameter values, and all remaining notations are defined as in equation~\ref{eq:HOP}. It is clear that in the very first query given to the root agent, we have: $\hat{\lambda}_o=\lambda_o\subseteq\lambda$ and $\hat{\lambda}'_o=\emptyset$.

\begin{algorithm}
	\DontPrintSemicolon
	\SetKwFunction{FStart}{\textsc{Start}}
	\SetKwFunction{FTune}{\textsc{Tune}}
	\caption{Distributed formation of the hierarchical agent-based hyper-parameter tuning structure, and the iterated hierarchical tuning of the hyper-parameters.}
	\label{alg:form-hier}
	\Fn{\FStart{$\left<\mathcal{A}_\lambda,\{\hat{\lambda}_o,\hat{\lambda}'_o, \lambda_f\},\mathcal{V}, \{\mathcal{X}^{(train)},\mathcal{X}^{(valid)}\}, \mathcal{L}\right>, c$}}{
		\eIf(\Comment*[f]{agent is terminal}){$|\hat{\lambda}_o|=1$}{
			$\mathcal{R}\gets$\Call{PrepareResources}{$\left<\{\hat{\lambda}_o,\hat{\lambda}'_o, \lambda_f\}, \{\mathcal{X}^{(train)},\mathcal{X}^{(valid)}\}\right>$}\;\label{ln:resource}
			\Call{Inform}{$Parent,\mathcal{R}$}\Comment*[f]{informs the parent agent}
		}
		(\Comment*[f]{agent is internal ($|\hat{\lambda}_o|>1$)}){
			$k\gets \min(c, |\hat{\lambda}_o|, \textnormal{MyBudget})$\Comment*[f]{the number of children}\\
			\For{$i\gets1$ \KwTo $k$}{
				$\mathcal{T}_i\gets$ \Call{SpawnOrConnect}{$\mathcal{A}_\lambda, \mathcal{L}$}\;\label{ln:spawn}
				$\hat{\lambda}_{o_i}\gets$\Call{Divide}{$\hat{\lambda}_o$,$i$, $k$}\Comment*[f]{the $i^{\text{th}}$ unique devision}\\
				$\hat{\lambda}'_{o_i}\gets(\hat{\lambda}_{o} - \hat{\lambda}_{o_i})\cup\hat{\lambda}'_{o}$\;
				$\mathcal{R}_i\gets$\Call{Ask}{$\mathcal{T}_i$, \textsc{Start}, $~~~~~~~~~~~~~~~\left<\mathcal{A}_\lambda,\{\hat{\lambda}_{o_i},\hat{\lambda}'_{o_i}, \lambda_f\}, \mathcal{V},\{\mathcal{X}^{(train)},\mathcal{X}^{(valid)}\}, \mathcal{L}\right>, c$}	
			}
			$\mathcal{R}\gets$\Call{Aggregate}{$\{\mathcal{R}_i\}_i$}\Comment*[f]{combines children's answers}\\
			\eIf{Parent $\ne\emptyset$}{
				\Call{Inform}{$Parent, \mathcal{R}$}\;
			}{
				$\mathcal{F}\gets\Call{PrepareFeedback}{\mathcal{R}, \mathcal{V}}$\;\label{ln:feed}
				\Call{Tune}{$\mathcal{F}$}\Comment*[f]{initiates the tuning process}\\
			}
		}
	
	}

	\Fn{\FTune{$\mathcal{F}$}}{
		\eIf{$Children=\emptyset$}{
			$\mathcal{R}^{(*)}\gets$\Call{RunTuningAlgorithm}{$\mathcal{F}$}\Comment*[f]{single-agent tuning}\label{ln:run}\\
			\Call{Inform}{$Parent,\mathcal{R}^{(*)}$}
		}{
			\ForEach{$\mathcal{T}_i\in{\normalfont Children}$}{
				$\mathcal{R}^{(*)}_i\gets$\Call{Ask}{$\mathcal{T}_i$, \textsc{Tune}, $\mathcal{F}_i$}\;
			}
			$\mathcal{R}^{(*)}\gets$\Call{AggregateResults}{$\{\mathcal{R}^{(*)}_i\}_i$}\Comment*[f]{combines results}\\
			\eIf(\Comment*[f]{non-root internal agent}){$Parent\ne\emptyset$}{
				\Call{Inform}{$Parent,\mathcal{R}^{(*)}$}\;
			}{
				\eIf{\textsc{ShouldStop}{\normalfont ($StopCriteria$)$\ne \text{True}$}}{\label{ln:stop}
					$\mathcal{F}\gets\Call{PrepareFeedback}{\mathcal{R}^{(*)}}$\;
					\Call{Tune}{$\mathcal{F}$}\Comment*[f]{initiates next tuning iteration}\\
				}{
					\Call{PrepareReport}{$\mathcal{R}^{(*)}$}\Comment*[f]{reports final result}
				}
			}
			
		}
	}
\end{algorithm}

The function and variable names in algorithms~\ref{alg:form-hier} are chosen to be self-explanatory, and additional comments are provided as needed. Additionally, some of the important parts are described as follows: function \textsc{PrepareResources} in line~\ref{ln:resource} prepares the agent, in terms of data and computation resources it will need for training, validating, and tuning the ML algorithm it represents; function \textsc{SpawnOrConnect} in line~\ref{ln:spawn} creates a subordinate agent, by either creating a new agent or connecting to an existing idle one, to represent algorithm $\mathcal{A}_\lambda$ and expected loss function $\mathcal{L}$; \textsc{PrepareFeedback} function in line~\ref{ln:feed} works on providing guidance to subordinate agents about their next move; and finally, function \textsc{ShouldStop} in line~\ref{ln:stop}, which is run by the root agent, determines when to stop based on a set of specified criteria, such as number of iterations, quality of improvements, etc. In the proposed model, the agents have flexible autonomy; at the same time that they are committed to follow the directions of their superior agents in the hierarchy and provide a specific format of responses, they can implement their own independent tuning algorithm. Moreover, using the feedback received from their parents, the terminal nodes have better control on handling defined and hidden constraints of the hyper-parameters.

There are various factors that determine the computational complexity of the entire process: (1) the tuning algorithm that each terminal agent uses and (2) the hierarchical structure of the MAS. Let $n$ denote the total number of hyper-parameters that we initially intend to tune and $t_i(n)$ be the temporal cost of the tuning algorithm run by terminal agent $\mathcal{T}_i$. Due to the recursive division of the objective hyper-parameters at each node of the hierarchy, and assuming that the parallelization budget of each agent equals $c$, it can be easily shown that the height of formed structure is $\lceil\log_c n\rceil$. Since the agents run in parallel, the formation of the structure has the worst case time complexity of $\mathcal{O}(\log_c n)$. Assume the maximum number of iterations before reaching the stopping criteria is denoted by $\mathcal{I}$, and the temporal cost of result aggregation process be $g(n)$. After building the hierarchy, the time complexity of running all aggregations for each iteration will be $\mathcal{O}(g(n)\log_c n)$, because of the parallel execution of the agents. Finally, recalling the fact that all actual tuning algorithms are conducted by the terminal agents and we have $n$ number of such agents in the structure, the time complexity of the model will be $\mathcal{O}(\mathcal{I}\times\max(g(n)\log_c n, t_1(n), t_2(n), \ldots, t_n(n)))=\mathcal{O}(\max(g(n)\log_c n, t_1(n), t_2(n), \ldots, t_n(n)))$. The space complexity, on the other hand, will depend on the number of agents created and the space complexity of tuning and aggregation algorithms. In the worst case, the hierarchical structure is full, and the total number of agents will be $\mathcal{N}=\sum_{j=1}^{\log_c n} c^{j}=\frac{1-nc}{1-c}$. Among all the agents, the internal ones will use a fixed amount of space for internal hourse keeping operations. Therefore, assuming $t'_i(n)$ and $g'(n)$ to be the space cost of terminal agent $\mathcal{T}_i$'s tuning algorithm and the result aggregation procedure, respectively, the time complexity of the proposed model will be $\mathcal{O}(\max(n.g'(n), t'_1(n), t'_2(n), \ldots, t'_n(n)))$.

\subsection{Guided Randomized Agent-Based Tuning Algorithm}
This section presents a multi-level randomized tuning technique -- we refer to it as Guided Randomized Agent-based Tuning (GRAT) -- not only to demonstrate how the generic model can be used in practice, but also to empirically analyze its performance and behavior.

Terminal agents' tuning strategy and internal agents' aggregation and feedback preparation techniques are the most important things to specify designing any tuning process based on the previously presented general model. The tuning strategy that we propose in this section is based on the randomized searching method that is widely used in practice. We have chosen this approach as the base strategy because of its broad usage as a baseline method in the literature and its natural parallelization capabilities. Furthermore, we assume that all terminal agents are homogeneous in the sense that they implement the same tuning algorithm. 

For the sake of clarity, we explain the used aggregation and feedback preparation operations before delving into the details of the used tuning strategy. According to line~\ref{ln:run} of algorithm~\ref{alg:form-hier}, each terminal agent returns optimal result it finds for each hyper-parameter. Let such result defined as $\mathcal{R}^{(*)}_i = \{(\lambda_j,\mathcal{V}^{(*)}_j, \Psi^{(*)}_j)\}$, where $\Psi^{(*)}_j$ denotes the response function value of sub-optimal hyper-parameter values $\mathcal{V}^{(*)}_j$ tuned by terminal agent $\mathcal{T}_j$. Receiving all the sub-optimal value sets from the subordinates, the inner agents of GRAT then aggregates the results by simply applying set union operator $\bigcup_{\mathcal{T}_i\in Children}\mathcal{R}^{(*)}_i $.

The bottom-up aggregation process continues until the results reach the root node, where the feedback for the next iteration is prepared. Let $\{(\lambda_i,\mathcal{V}^{(*)}_i, \Psi^{(*)}_i);\\ 1\leq i\leq n\}$ be the set of all results merged and received by the root node. For each objective hyper-parameter $\lambda_i\in\hat{\lambda}_o$ in a minimization task, the root agent assigns the sub-optimal values corresponding the minimum response function value, $\Psi^{(*)}_j$, of all the ones found by the terminal agents representing other hyper-parameters, $j\ne i$. That is: 
\begin{equation}
	{\textsc{PrepareFeedback}}(\{(\lambda_i,\mathcal{V}^{(*)}_i, \Psi^{(*)}_i)\})=\left\{(\lambda_i, \mathcal{V}_j); j = \argmin_{1\leq j\ne i\leq n} \Psi^{(*)}_j\right\}
\end{equation}

The feedback will be split and propagated recursively until it reaches the terminal agents. Let $\mathcal{F}_i=\mathcal{V}_j$ be the feedback received by agent $\mathcal{T}_i$. Starting from the location specified by $\mathcal{F}_i$, the agent follows the steps in algorithm~\ref{alg:strategy} to locate a hyper-parameter values that yield $\Psi(\lambda)$ that is relatively lower than the one for the starting values. To do the search, GRAT divides the domain of real-value  $\hat{\lambda}_o=\lambda_i$ into a pre-specified number of slots, $\eta_i$, and uniformly chooses a value in each. For the remaining hyper-parameters, it chooses either the same value specified by $\mathcal{F}_i$ or a new one depending on a weight factor, $\omega_i\in\mathbb{Z}^+$, and whether the hyper-parameter belongs to $\hat{\lambda}'_o$ or $\lambda_{f}$.  

\begin{figure}
	\centering
	\includegraphics[]{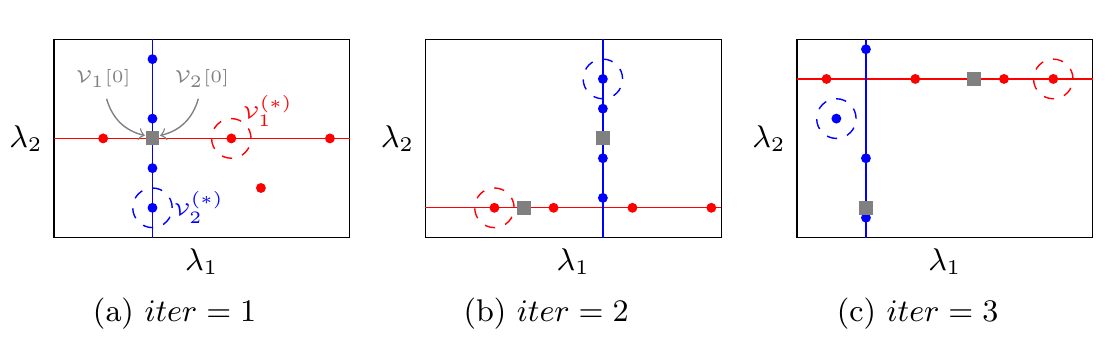}
	\caption{A toy example demonstrating 3 iteration of running GRAT on two hyper-parameters $\lambda_1$, and $\lambda_2$ using red and blue terminal agents respectively. In this example, we have assumed $\eta=4$.}
	\label{fig:tune-ex}
\end{figure}

\begin{algorithm}
	\DontPrintSemicolon
	\SetKwFunction{FRTune}{\textsc{RunTuningAlgorithm}}
	\caption{Guided Randomized Agent-based Tuning (GRAT) process.}
	\label{alg:strategy}
	\Fn{\FRTune{$\mathcal{F}=\mathcal{V}_i$}}{
		$\mathcal{V}[0]\gets \mathcal{V}_i$\;
		\For{$s\gets 1$ \KwTo $s=\eta$}{
			$\mathcal{V}[s](\lambda_i)\gets$\Call{UniformRand}{$\lambda_i, \eta, s$}\label{ln:uniform}\;
			\ForAll{$\lambda_k\in\lambda - \lambda_i$}{
				\eIf{$\lambda_k\in\hat{\lambda}_{o}$}{
					$\mathcal{V}[s](\lambda_k)\gets$\Call{WeightedRand}{$\lambda_{k}, \omega$}\label{ln:weight}\;
				}{
						$\mathcal{V}[s](\lambda_k)\gets\mathcal{V}(\lambda_f)$\;
				}
			}
		}
		$\mathcal{V}^{(*)}\gets\argmin\limits_{0\leq j\leq\eta} \Psi(\mathcal{V}[j])$\;
		\Return $\{(\lambda_i, \mathcal{V}^{(*)}, \Psi(\mathcal{V}^{(*)})\}$\;
	}
\end{algorithm}
Similar to the previous algorithms, the functions' and variables' names are chosen to be self-explanatory. Function \textsc{UniformRand} in line~\ref{ln:uniform} chooses a random value in the $s^{\text{th}}$ slot based on the uniform distribution. In case the domain of $\lambda_i$ is discrete, eg. in nominal type hyper-parameters, the function samples the values without replacement. In our implementation for function \textsc{WeightedRand} in line~\ref{ln:weight}, we assume similar $\eta$ slots for the real type hyper-parameters and choose the current value $\mathcal{V}[0](\lambda_k)$ with weight $\omega$ or a value in other slots with weight 1. In such formulation, the probabilities of choosing the value of the feedback set and a random value in any other slots will be $\frac{\omega}{\omega+(\eta - 1)}$ and $\frac{1}{\omega+(\eta - 1)}$ respectively. Depending on the improvements, the root agent might adjusts the weight through iterations to induce exploration in agents' searching process. Figure~\ref{fig:tune-ex} provides a toy example to briefly demonstrate 3 iterations of GRAT.  

\section{Results and Discussion}\label{sec:results}
This section presents the empirical results of using GRAT for both hyper-parameter tuning and function optimization use cases. In all of the experiments, we have assumed that $c=2$ and agents' $budget=\infty$. We have assessed the behavior of our model based on different metrics and compared the results with the ones from normal and latine hyper-cube randomized methods. To make the comparisons fair, in all experimental settings, we have fixed the total number of evaluations for both randomized methods to $c\times\eta\times\mathcal{I}$. All experiments have been carried out on a desktop with Intel Core-i5 @ 1.6GHz CPU and 16 GB RAM running Ubuntu 20.04 OS and Python 3.7. Additionally, all the used ML resources are from scikit-learn library~\cite{scikit-learn}, and the results reported in both ML and function optimization experiments are averaged over 100 trials.

\begin{figure}
	\centering
	\includegraphics[]{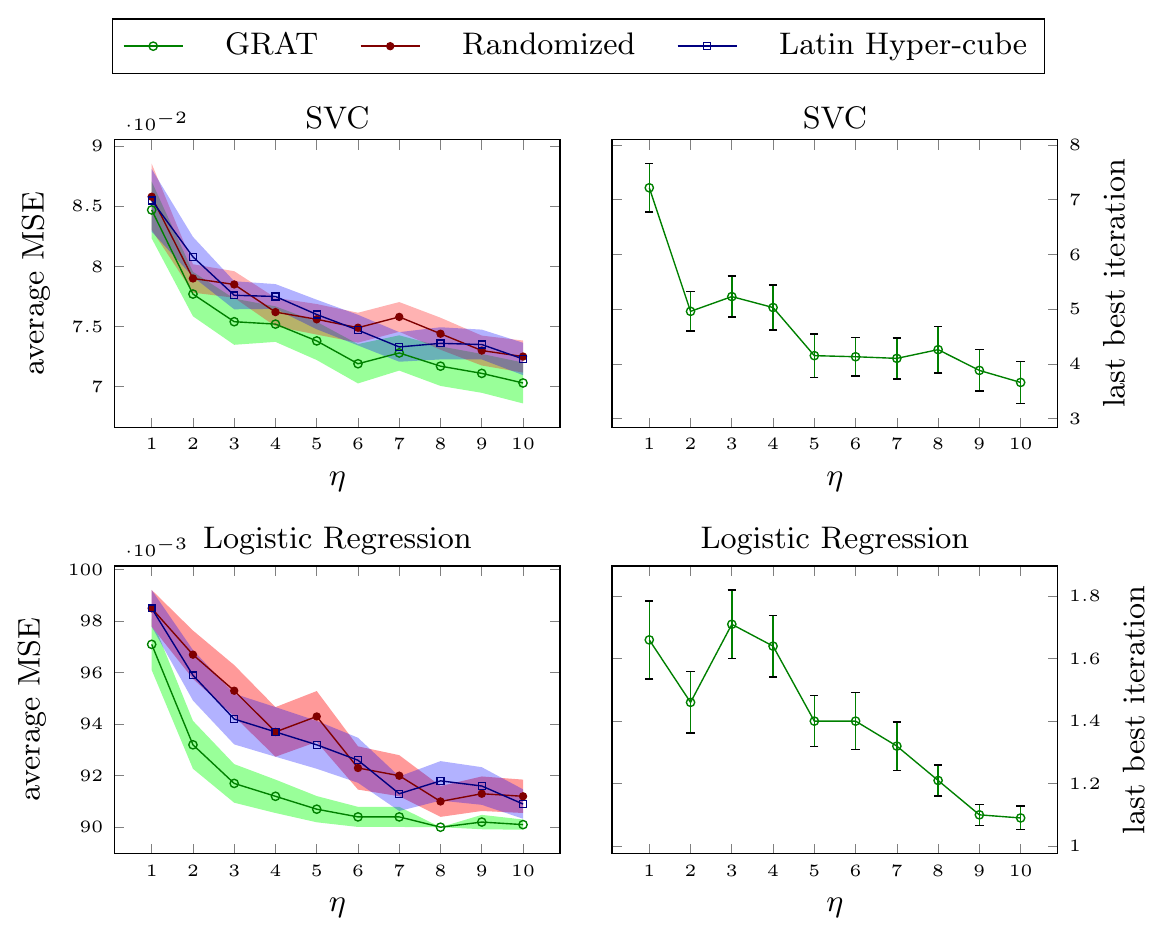}
	\caption{Average performance of SVC (first row) and Logistic Regression (second row) classifiers on two synthetic classification datasets based on mean squared error and the last best iteration number. The color bands on the left column plots represent 95\% confidence interval, and the error bars on the right column plots are calculated based on the standard error.}
	\label{fig:mlopex}
\end{figure}

In ML hyper-parameter tuning use case, we have chosen C-Support Vector Classification(SVC)~\cite{chang2011libsvm} and Logistic Regression (LR) algorithms on two synthetic datasets of size 100 with 2 classes and 20 features, both generated by scikit-learn's make\_classification tool~\cite{scikit-learn}. The ML experiments focuses on studying the behavior of GRAT at different number of randomized points that terminal agents are allowed to evaluate, i.e. $\eta$. For SVC we have set $\lambda_1=C\sim logUniform(10^{-2},10^{13}), \lambda_2=\gamma\sim Uniform(0,1), \lambda_3=kernel\in\{poly,linear,\\ rbf, sigmoid\}, \mathcal{I}=15$. Similarly, for LR we have $\lambda_1=C\sim logUniform(10^{-2},\\10^{13}),\lambda_2=solver\in\{newtoncg,linear, lbfgs, liblinear\}, \mathcal{I}=5$, where $C$ and $solver$ are respectively the inverse of regularization strength and LR's  optimization algorithm. In all of the experiments, we have used 5-fold cross validation with stratified sampling as the model evaluation method. Figure~\ref{fig:mlopex} depicts the results obtained in each classification task, along side the average value of the last best iteration count -- any increase in the iteration count beyond this value has not improved the performance. In SVC, we have used mean squared error (MSE) to measure the performance and as it can be seen, GRAT has outperformed the other two methods in almost all $\eta$ values, while the model evaluation counts for all methods is the same. Furthermore, with an increase in $\eta$, we observe a decrease the number of iterations needed to find the improved performance. LR also uses MSE as the performance metric, and we have a very similar behavior for GRAT, together with the point that the 95\% confidence intervals are narrower, and the improved values are obtained with lower values for $\eta$.

\begin{figure}
	\centering
	\includegraphics[]{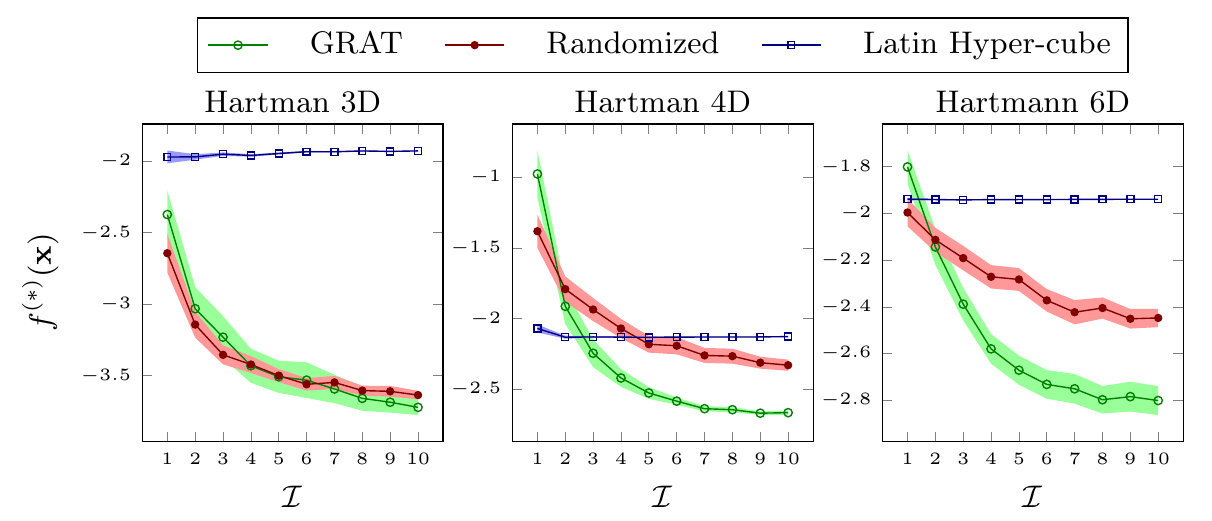}
	\caption{Optimal function value, ${f^{(*)}}(\textbf{x})$, vs iteration threshold $\mathcal{I}$, averaged over 100 different experiments. The plots include 95\% confidence interval bands of the results.}
	\label{fig:funopex}
\end{figure}

For function optimization, we have used three multi-modal benchmark functions widely used in the optimization literature: Hartman-3d, Hartman-4d, and Hartman-6D~\cite{Jamil_2013}. In these sets of experiments, we mainly focus on the impact of increasing the iteration threshold and the size of dimensions on the performance. Figure~\ref{fig:funopex} illustrates the average function values found at different iteration thresholds. In all of the experiments, we have fixed the value of $\eta$ to 10. As it can be seen in , the average function values that have been found by GRAT are less (better) than the other two methods -- this is a minimization problem. Furthermore, a careful look at the plots reveal that the improvements of the results are significantly more when the number of dimension, i.e. the number of the parameters that we optimize, increases from 3 to 4 and then to 6.

It is worth noting that the goal in these sets of experiments are not to compete with the state-of-art hyper-parameter tuning and function optimization methods but to demonstrate how our proposed model can be used to develop distributed tuning/optimization techniques and how effective it is in comparison to the baseline methods. We believe our core model is capable of providing competitive results provided with more sophisticated and carefully chosen tuning strategies and corresponding configurations.   

\section{Conclusion}\label{sec:conclusion}
In this paper we presented a hierarchical multi-agent based model for machine learning hyper-parameter tuning and function optimization. The presented model includes a two-phase procedure to distributively form the hierarchical structure and collaboratively tune/optimize the hyper-parameters. Using its flexible configuration, our proposed generic approach supports the development of solutions that comprise diverse sets of tuning strategies, computational resources, and cooperation techniques, and we successfully showed how it can be used in practice for both ML hyper-parameter tuning and global function optimization tasks. 
For the sake of evaluation, this paper presented analytical discussion on the computational complexities and implemented a basic guided randomized agent based tuning technique, called GRAT. The empirical results, obtained from both ML classification and multi-modal function optimization use cases, showed the success of the suggested collaborative model in improving the performance of the underlying tuning strategies, especially in higher dimension problems.  

This paper provided the foundations and a basic implementation for a novel agent-based distributed hyper-parameter tuning model. Despite its demonstrated success in small size problems, the research can be extended in various directions such as investigating its behavior and performance in large scaled deep learning based structures, conducting a comprehensive sets of analysis on the impact of architectural and strategic diversity on the performance of the model, and employing more sophisticated and data-efficient approaches such Bayesian optimization as the optimization strategies of terminal agents.. We are currently working on some of these studies and suggest them as future work.
%
%
%
 \bibliographystyle{splncs04}
 \bibliography{refs}

\end{document}